\documentclass[10pt,twocolumn,letterpaper]{article}

\usepackage{cvpr}
\usepackage{times}
\usepackage{epsfig}
\usepackage{graphicx}
\usepackage{amsmath}
\usepackage{amssymb}

\usepackage[breaklinks=true,bookmarks=false]{hyperref}

\cvprfinalcopy 

\def\cvprPaperID{****} 

\begin{document}


\title{Deep Learning Methods for Efficient Large Scale Video Labeling}

\author{Miha Skalic\footnotemark[1] \\
University Pompeu Fabra, Barcelona\\
{\tt\small miha.skalic@upf.edu}
\and
\\
Marcin Pekalski\footnotemark[1] \\
{\tt\small mpekalski@gmail.com}
\and
\\
Xingguo E. Pan\footnotemark[1] \\
{\tt\small panxingguo@gmail.com}
}

\maketitle
\let\thefootnote\relax\footnotetext{*These authors contributed equally to this work}

\begin{abstract}
   We present a solution to ``Google Cloud and YouTube-8M Video Understanding Challenge'' that ranked 5th place. The proposed model is an ensemble of three model families, two frame level and one video level. The training was performed on augmented dataset, with cross validation.
\end{abstract}

\section{Introduction}
Gathering huge datasets and organizing competitions \cite{ILSVRC15} \cite{DBLP:journals/corr/LinMBHPRDZ14} has driven forward machine learning model
development. In fact, some of the key breakthroughs were achieved as part of these competitions \cite{NIPS2012_4824} \cite{DBLP:journals/corr/VinyalsTBE14}. However, large scale video predicting remains to date a non-trivial problem. To tackle this problem Google Cloud has hosted a Kaggle competition with a dataset consisting of 6.3 million and 740 thousand labeled and unlabeled  videos, respectively. The labels vocabulary consisted of 4716 labels. Participants were asked to develop a machine learning model that would predict what labels were assigned to each of the videos from the test dataset.

\subsection{Video Data and Competition Starter Code}
The organizers provided two datasets; one on video level features and another on frame level features. The frame level consisted of 1024 RGB features and 128 audio features for each frame of the video sampled every second up to 300 frames. Those RGB features were extracted from the last ReLu activation of the hidden layer, just before the classification layer (layer named pool3/\_reshape) of the Inception-v3 network. After extraction, these RGB features were dimension reduced by PCA and whitened. The video level features are simply an average of the PCA and whitened frame level features. For more details please refer the original YouTube-8M dataset paper \cite{DBLP:journals/corr/Abu-El-HaijaKLN16}.
The dataset is available on the project's website, https://research.google.com/youtube8m/.

In addition to supplying the video data, the dataset authors also provided a starter code base for the competition. The starter code is well structured and provided the whole pipeline from data input, model training to prediction output. It included several base models and proved to be a robust baseline for further creative model work.

\subsection{Evaluation}
The  Global Average Precision was picked as evaluation metric. It is defined as defined as
$$GAP=\sum\limits_{i=1}^{N}p(i)\Delta r(i).$$
Where $N$ is the number of final predictions (if there are $20$ predictions for each video, then $N=20*\mathbin{\#}Videos$), $p(i)$ is the precision score, and $r(i)$ is the recall score.

In the following three sections we describe key elements of our work. First we describe data augmentation, followed by descriptions of used models and their assembly. Finally, we describe model training techniques that we tested and used. Our final solution includes both frame level and video level models. Hence, the further description with cover both cases, but not everything is applicable for all models.

\section{Data}

\subsection{Video Level Feature Extraction}

The provided video level data consists of $mean\_rgb$ and $mean\_audio$. Google Research team reported better model performance by adding second moments and top 5 ordinal statistics of each feature dimension \cite{DBLP:journals/corr/Abu-El-HaijaKLN16}.

Our feature engineering efforts started with similar extensions: 
we first generated first three moments of $rgb$ and $audio$ features from the frame level data. We named them $mean\_rgb$, $mean\_audio$, $std\_rgb$, $std\_audio$, $x3\_rgb$ and $x3\_audio$. Than we also used the number of frames of each video, $num\_frames$, and the second moments of $rgb$ and $audio$ of the entire frame level video data. All these features were used in some combination in our video level models, and all of them contributed to the model performance. Having noticed high correlation between [$mean\_rgb$, $mean\_audio$] and [$x3\_rgb$, $x3\_audio$], for a given model we will only include one of the two groups. 

Trying to capture some non-linearity we also introduced some basic transformations of $mean\_rgb$ and $mean\_audio$, those included taking log, inverse, and square root of the input. Due to the lack of time and computing power, we did not experimented with these nonlinear features. Google's research paper mentioned added value from top-5 features. We experimented with them in the early stage but did not extract much value, as a result we did not use them in our final models.

\subsection{Training Data Augmentation}\label{ss:augmentation}

Getting more training data is one of the most effective ways to improve model performance. We took two approaches to achieve this goal. First, we decided to integrate validation set with the training data set. There are 4.9 million videos in the training set and 1.4 million in the validation set. The integrated training data is 29\% larger. When training a model, we split the data in the same fashion as a five fold cross-validation, and trained one set of model variables on each of the 80\% fold. This allowed us to monitor on-line evaluation GAP and also to ensemble the predictions from the five folds. As reported in the next section, we saw material improvement of the model performance from this fold setup.

The second approach we took to augment the training data is by splitting the videos. Our rationale is that, since a person watching a 5 minute video should be able to tell its topic after seeing either of the halves, the model should be able to do the same. By this approach, each of the frame level video yielded three sets of video level data: one from the entire video, one from the first half, and one from the second half. Therefore in total we had about 19 video examples to use in training. It is worth mentioning that the three data sets from the same video are correlated at a high level. Still the added value of the augmented data was clearly visible. Naturally one would think of splitting the video into even more parts. We investigated dividing them into 3 parts, but we did not observe additional performance gain over 2 parts.

When going through the video data we noticed that about 0.1\% of the frame level data had only one frame. One of our team member reported this to competition organizers. We believe there will be a new version released for these videos after the competition has finished.

\section{Models}

The final model is a weighted ensemble of ensembles of Mixture of Neural-Network Experts (MoNN), Long Short-Term Memory (LSTM) \cite{Hochreiter1997} and Gated Recurrent Units (GRU) \cite{DBLP:journals/corr/ChoMGBSB14}. In this section we will describe what hides behind those abbreviations and provide some more details of which of the mentioned techniques we used. In the last subsection we will briefly write about what we have tried that did not work as well as expected.

\subsection{MoNN}
Our major video level model is called MoNN, Mixture of Neural-Network Experts. It is a Mixture of Experts (MoE) model with inter-connected multi-hidden layer neural networks. The MoE model, first introduced in \cite{Jacobs:1991:AML:1351011.1351018}, consists of two parts gating and experts. The former ends with a softmax layer and the latter with sigmoid. Then the final predictions are a sum of products of last layers from gating and experts. In our extension to the model we tried adding more layers for both gating and expert parts. Our best preforming model is called MoNN3Lw. It had 3 hidden layers in the expert part of the network (3L), and the layers are relatively wide (denoted by "w" in the name). Their sizes were $2305*8$, $2035$, and $2305*3$ neurons. Gate activation were left as they were set in the sample code with first layer of $4716 \times ({\rm num\ of\ experts} + 1)$. Our submitted predictions were mostly run with three experts, a balance between model size and the GPU memory constraints. We also calculated the exponentially weighted moving average (EMA) for the model variables (weights and biases) with a half-life set to 3000 or a smaller number of steps. In the prediction stage, we can choose to use the either the EMA or the snapshot version of the model.

For each model training session, the augmented training data were then split into folds like in a classical cross validation approach. For a model defined as above we run 5 folds, saving checkpoints every 6 K steps up till 72 K steps (about 5 epochs). Then we picked one fold and for each of last 3 to 5 checkpoints and made predictions. We calculated predictions with either the EMA model variables or the snapshot version of variables. We found that the EMA model consistently scored higher, so we stayed with EMA most of the time.

One example of our MoNN3Lw ensemble is shown in Table. \ref{table:MoNN3Lw}. This model was trained on a Nvidia GeForce GTX 1080Ti GPU. It took around 10 hours to run one fold, and making predictions for all 5 checkpoints. The typical single fold and single checkpoint prediction scored GAP 0.822. Ensemble of 4 checkpoints from the same fold scored 0.824. Ensemble of 5 folds each with 4 checkpoints scored 0.831. In total we saw a GAP gain close to 0.01 from a single prediction to model ensemble.

\begin{table}
\begin{center}
\begin{tabular}{|c|c|c|}
\hline
Used in prediction & Public LB & Private LB \\
\hline\hline
ckpt 36050 & 0.82126 & 0.82135 \\
ckpt 48050 & 0.82297 & 0.82294 \\
ckpt 54050 & 0.82274 & 0.82265 \\
ckpt 60050 & 0.82216 & 0.82209 \\
ckpt 66050 & 0.82034 & 0.82016 \\
\hline
\shortstack{ensemble one fold \\ ckpts
36k48k54k60k} & 0.82411 & 0.82408 \\
\shortstack{new ensemble 5 folds \\ ckpts 36k48k54k60k } & 0.83093 & 0.83087 \\
\hline
\end{tabular}
\end{center}
\caption{3 layer MoNN model with EMA, leaderboard scores.}
\label{table:MoNN3Lw}
\end{table}

We experimented further with EMA parameter, and found that it gives slightly better results when it is shorter than 3k steps. We have included that in another model that instead of means took into account third powers of each dimension. We hoped that it will be slightly less correlated to the model based on means than just yer another fold of model based on means. 

Beside the two models mentioned above we did similar exercise for a couple more models. The final ensemble for MoNN type models consisted of the models listed below, where $w$ in model's name stands for wide, $n$ for narrow, "means" represents $[mean\_rgb, mean\_audio]$, and "third powers" represents $[x3\_rgb, x3\_audio]$. $std\_rgb$, $std\_audio$ and $num\_frames$ are always included in the training.

\begin{enumerate}
  \item MoNN3Lw, based on means, with 3 layers \\ $2305*8$, $2035$, and $2305*3$, no EMA;
  \item MoNN3Lw, based on means, with 3 layers \\ $2305*8$, $2035$, and $2305*3$, with EMA;
  \item MoNN3Lw, based on third powers, with 3 layers \\ $2305*8$, $2035$, and $2305*3$, with EMA;
  \item MoNN2Lw, based on means, with 2 layers \\ $2305*6$ and $2305*3$, no EMA;
  \item MoNN2Lw, based on means, with 2 layers \\ $2305*6$ and $2305*3$, with EMA;
  \item MoNN2Lw, based on third powers, with 2 layers \\ $2305*6$ and $2305*3$, with EMA;
  \item MoNN3L, based on means, with 3 layers \\ $4096$, $4096$, and $4096$, no EMA;
  \item MoNN4Ln, based on means, with 4 layers \\ $2048$, $2048$, $2048$, and $2048$, no EMA;
\end{enumerate}

To summarize, we made many models of MoNN class that are a mix of different features and slightly different structures (2L, 3L, 4L, narrow and wide, with or without EMA). We found that equal weights ensemble worked well for this class of models.

\subsection{LSTM}
In contrast to video level models, frame level models turned out to be more time expensive to train. 6 epochs of LSTM model training took approximately 3 days to train on Nvidia GTX 1080ti GPU. Consequently we prioritized exploitation of good models over exploration of new ones. In our final submission we used two types of recurrent neural networks: LSTMs and GRUs. 

\begin{figure}
\begin{center}
\fbox{
\includegraphics[width=0.9\linewidth]{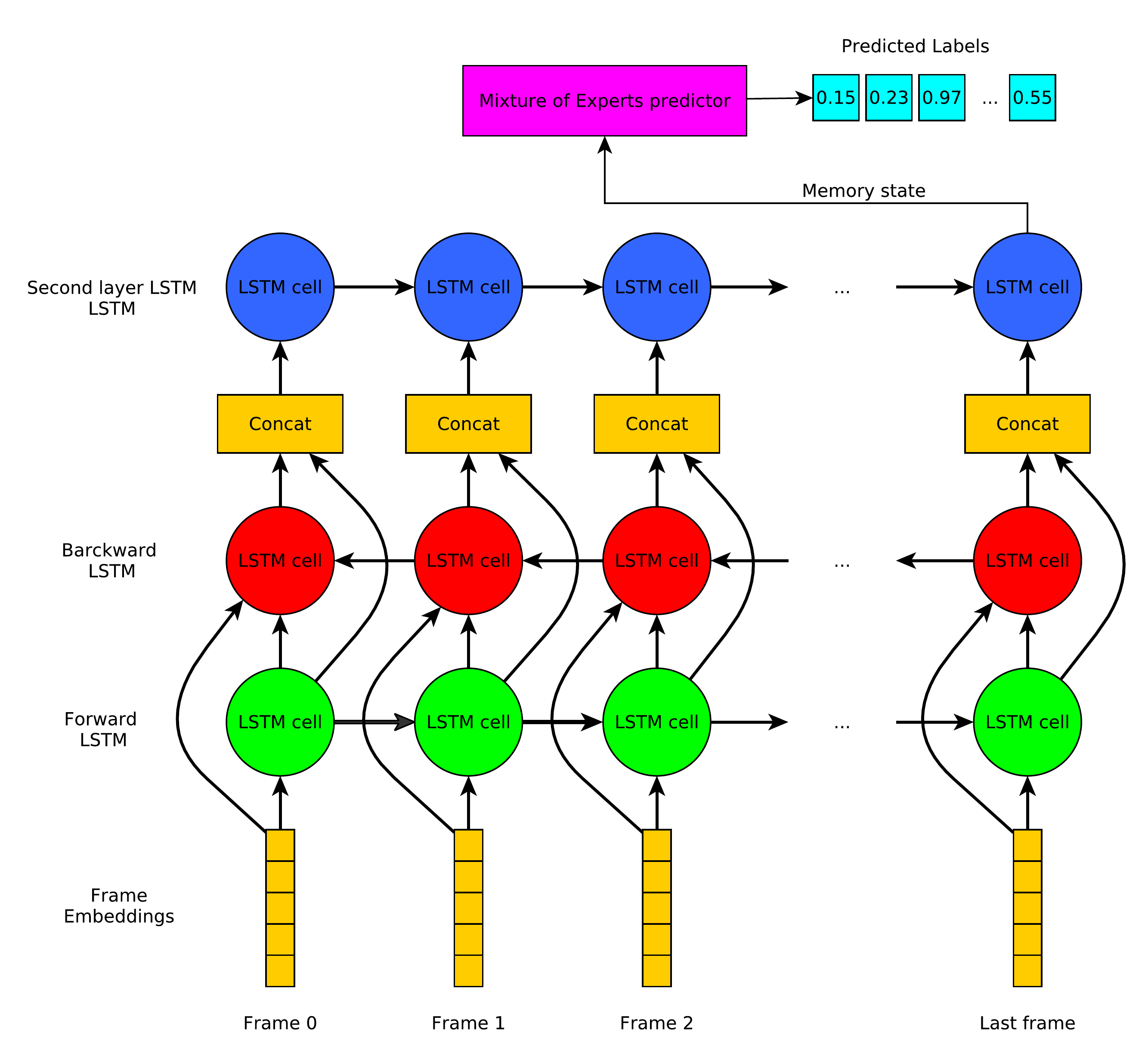}
}%
\end{center}
   \caption{LSTM architecture.}
\label{fig:GLSTM}
\end{figure}

For both types of recurrent networks we used dynamic (varied length) versions of Recurrent Neural Networks (RNNs) with maximum length of 300 frames. We propagated the input through 2 layer network. The first layer was a bidirectional RNN, a network consisting of two models running in the opposite direction. Outputs of the two models at each step were then concatenated before being feed to the second single direction layer. Memory of the last state was used as input to a MoE network with two experts. Schematic representation of the networks presented in Fig. \ref{fig:GLSTM}. In case of LSTMs we used 600 and 1200 unit cells for bidirectional and single direction RNNs, respectively. Models were trained with an initial learning rate of 0.00025 and batch size of $256$. We used the default starter code values for all other hyperparameters. In total we trained 3 models from scratch. We trained on all the available data. We were able to boost the GAP score by: 
\begin{enumerate}
  \item Assembling predictions of 3 checkpoints.
  \item Adding models that were trained using different initialization.
  \item Continue training the model while keeping EMA values of the model weights and using the EMA model variables for prediction.
\end{enumerate}
All three techniques helped improve our GAP score (Table \ref{table:LSTM}).
Our final submission consisted of model predictions after training for approximately 115, 125 and 135 thousand batches. Additional training with EMA of the weights was done for 3000 batches with a half life of 3000 steps.

\begin{table}
\begin{center}
\begin{tabular}{|l|c|c|}
\hline
Model & Public LB & Private LB \\
\hline\hline
1 model, 1 checkpoint & 0.81547 & 0.81592 \\
1 model, 3 checkpoints & 0.82244 & 0.82276 \\
2 models 3 checkpoints & 0.83085 & 0.83100 \\
1 model, 1 checkpoint, EMA & 0.81978 & 0.82012 \\
2 models 3 checkpoints, EMA & 0.83183 & 0.83183 \\
\hline
\shortstack{final ensemble \\ 3 model, 3 checkpoints, EMA } & 0.83482 & 0.83489 \\
\hline
\end{tabular}
\end{center}
\caption{LSTM predictions, leaderboard scores. A small score variation is present between the models and within model checkpoints.}
\label{table:LSTM}
\end{table}

\subsection{GRU}

For the GRU models we used similar architecture as for the LSTM. The only exception was that we used 1250 (second layer) and 625 (first layer forward and backward) unit cells. Training time of about 3 days for 6 epocs was comparable to LSTM training time. Unlike LSTMs, we trained GRUs using bootstrap aggregating. We used 80\% of the data for training and leaving out the remaining 20\%. We did this for 5 folds, in each fold leaving out different set of samples. Final evaluation was done on 5 checkpoints after training for approximately 82, 94, 96, 98 and 100 thousand batches.

\subsection{Other Frame Level Models}

In addition to the presented LSTMs and GRUs we also trained several other frame level models. Although we did not use these models in the final ensemble we hypothesize that they might be useful if properly optimized and assembled. We made these models available as part of our code. 

We tried to increase the width of LSTM models by feeding every second frame into a different 2 layer LSTM network. Last memory states were then summed up before being feed into the MoE classifier. A 5 fold bootstrap aggregating submission with 3 checkpoints yielded private and public GAP score of 0.82734 and 0.82744, respectively. This model gave better GAP score then the starter code LSTM network, but lower score than previously described bidirectional model.

Next, we tried coupling LSTM aggregation with regularized MoNN classifier instead of MoE classifier. When training a model with a single hidden layer we observed that the training was substantially slower. A 7 checkpoint prediction (evenly spaced checkpoints in range 160k-210k; 128 samples batch size) achieved private leaderboard GAP score of 0.81523, which is significantly lower than bidirectional 3 checkpoints model.

When testing DBoF models we saw that making the pooled layer wider helps improving the GAP score. We made a 5 fold bootstrap aggregating DBoF ensemble with average pooling and 11 thousand units layer before the pooling operation. A 5 folds and 6 checkpoint ensemble yielded private leaderboard score of 0.82144. Unfortunately, when using these predictions together with RNNs and MoNNs we did not see improvement of the score. 

We also tried training a single hybrid DBoF/GRU network. In this network we concatenated the aggregated values before feeding them into final 2 expert MoE classifier. However, we were not able to fully train this network before the competition ended. We observed the network kept good properties from both networks. the training GAP score quickly increased (as observed in default DBoF networks) and it did not plateaued after several thousand batches (as observed in RNN networks).

Finally, we explored usage of 1D convolution as an alternative frame aggregating method. In our initial experiment where we combined convolutions with different filter lengths, strides, frame skipping and polling methods our training GAP score plateaued at $\sim0.72$. We believe that increasing the model capacity would improve the score.


\begin{table}
\begin{center}
\begin{tabular}{|l|c|c|}
\hline
Model & Private LB & Public LB \\
\hline\hline
1 fold, 1 checkpoint & 0.81200 & 0.81197 \\
1 fold, 3 checkpoints  & 0.81759 & 0.81766 \\
5 folds, 1 checkpoint  &  0.83143 & 0.83156  \\
5 folds, 3 checkpoints  &  0.83306 & 0.83322 \\
\hline
\shortstack{final ensemble \\ 5 folds, 5 checkpoints} & 0.83349 & 0.83362 \\
\hline
\end{tabular}
\end{center}
\caption{GRU predictions, leaderboard scores.}
\label{table:GRU}
\end{table}
\subsection{Model Ensemble}
For each of our major models, except for LSTM model, we trained them on five data folds. For each fold, we again picked about five checkpoints that are about half an epoch apart in the fully trained regime. We then calculated the equal weighted average probability of the 25 predictions. For the video level model MoNNs, the ensemble prediction GAP is generally 0.01 higher than the single fold and single checkpoint prediction GAP. For the frame level models, the gain is even higher at 0.02. 

As expected, ensembles from model predictions with low correlation tend to enjoy higher added value. For effective model ensemble, we created a measure of prediction correlation. For one video, we viewed a prediction as a 4716-dimension vector. Following the standard approach in vector analysis, we define the correlation between two predictions as:
\[ \rho = \frac{{\mathbf p_1} \cdot 
{\mathbf p_2}}{|{\mathbf p_1} | \;|{\mathbf p_2}|} \ ,\]
where ${\mathbf p_1}$ and ${\mathbf p_2}$ are two predictions and each is a $R^{4716}$ vector of probabilities. In practice, we only kept the first 20 to 120 dimensions when we calculated correlations  or merged predictions. We believe the measure is robust and reflected the relations among the models well.

We considered and experimented with several ensemble approaches including Kelly strategy \cite{6771227} or mean-variance type optimization. In the end due to tight computing resources we took practical approach and discretionarily determined the weights of component models based on both the single model GAP and the correlation matrix.

Our final prediction is a weight summation of the three model families: MoNNs, LSTMs, and GRUs. The inidivual components' scores and their correlation matrix are shown in Table. \ref{table:final_ensemble_score} and Table. \ref{table:final_ensemble_corr}. And the weights were 0.40, 0.36, and 0.24 for MoNNs, LSTMs, and GRUs, respectively. Our best score was 0.84185 on Private Leaderboard and 0.84192 on Public Leaderboard. The ensemble gained 0.007 from the best individual component model.

\begin{table}
\begin{center}
\begin{tabular}{|l|r|r|}
\hline
Model & Private LB & Public LB \\
\hline\hline
MoNNs & 0.83496 & 0.83484 \\
LSTMs & 0.83482 & 0.83489 \\
GRUs &  0.83349 & 0.83362 \\
\hline
\end{tabular}
\end{center}
\caption{Final ensemble component score}
\label{table:final_ensemble_score}
\end{table}

\begin{table}
\begin{center}
\begin{tabular}{|l|l|l|l|}
\hline
Model & MoNNs & LSTMs & GRUs \\
\hline\hline
MoNNs &  1.0 & 0.9588 & 0.9611 \\
LSTMs &      & 1.0    & 0.9772 \\
GRUs &       &        &   1.0     \\
\hline
\end{tabular}

\end{center}
\caption{Final ensemble component correlation}
\label{table:final_ensemble_corr}
\end{table}


\section{Model Training Techniques}

The baseline models provided by the organizers were much more than what has ever been seen on Kaggle. It included basic implementations of Logistic Regression, Mixture of Experts (MoE), Long Short Term Memory (LSTM) and Deep Bag of Frames (DBoF). Each of them has many parameters to pick or tune. The code was well structured for further experimentation. In this section we will briefly describe several ideas we tried and state if they worked for us or not.

\subsection{Run-time Validation}
When training video level models, one of the most helpful techniques we have introduced was the run-time validation. 
Every 10-50 steps we run the evaluation of currently trained model on a holdout sample. In the beginning we run it on the validation sample prepared by the organizers, but after we included it in the training dataset we run it on a randomly selected holdout subset. This allowed us to clearly see the in-sample and out-of-sample performance of the training, and allowed us to stop the training when it has entered the over-fitting regime. For MoNN models, we found that the best out-of-sample performance was achieved between 2.5 and 4 epochs over the augmented data set.

\subsection{Dropout}
The very first novelty in the code we introduced was dropout layers, originally described in \cite{2012arXiv1207.0580H}. The ides behind it is to randomly drop some neurons during training phase to prevent over-fitting but during inference use all of them. Introducing dropout extends learning time significantly and adds a couple more parameters to train. For this data set, we saw mixed performance from dropout layers. Our final models did not use dropout.

\subsection{Truncated Labels}
The vocabulary of labels consisted of 4716 expressions. Some of them were really popular, like Games (865k), Vehicle (683k), Video Game (522k), Car (373k), and some really rare Russian Pyramid (102), or PBS Kids (101). We decided to explore models that would work on a smaller vocabulary, i.e. excluding bottom 1k labels. We hoped that by focusing on a small subset of labels, we could avoid variances introduced by less popular labels. The best single model prediction, which fitted only the first 3100 labels, achieved a score of 0.80082 on the Public Leaderboard without any ensemble or EMA. Although the correlation between truncated model prediction and the regular predictions is low (e.g., 0.92\%), the truncation predictions did not add value to ensemble due to their lower scores. We did not use any of them in our final models.

\subsection{Exponential Moving Average}
For both video level and frame level models, we calculated the exponential moving average of model variables. And in the prediction phase, we can choose either the EMA or the snapshot variable values to perform the inference. We found that, for single checkpoint, using EMA we were able to increase the MoNN model GAP score by 0.006 and LSTM model by 0.004. However, the ensembles of the EMA based predictions only improved by 0.0015 for MoNN family. We think this indicates that the variance reductions by model ensemble and by EMA have overlapping components. Still, EMA concretely helped to improve the overall model performance.

\subsection{Batch normalization and global normalization}

When training the video level models, we experimented with batch L2 normalization, global L2 normalization, and in the end found our models performed better without any L2 normalization. Batch normalization is expected to make the training converge sooner, however at the same time it introduces variation as the values for one observation may depend on what other observations happened to be in the same batch. We observed that this data set is highly homogeneous and decided to calculate the global moments of each feature dimension based on the train data. We then applied a global L2 normalization using the calculated train data moments. The model converged as fast as the L2 normalization case and reached a slightly higher validation GAP. Later on we found that the models performed even better if no L2 normalization was applied. We believe that the relative value among the different dimensions of the data carried information. This reason may come from the way the data were prepared. The PCA process naturally lined up the components with decreasing variance and the relative variance carried information. An L2 normalization step would remove this information.

\subsection{Compression}
Although, Google was very generous to give some Google Cloud ML credits for competition's participants, they were only enough to test a couple of big models. Most of the work was still done locally. One of the challenges was to feed data as fast as possible to the GPU, to keep it fully utilized. One of the ways to increase I/O is data compression. As CPU is barely utilized when you run models on a GPU it can be used to decompress the data without any performance hits. Also compressed files are smaller so using them should result in faster reading speeds from HDD (due to the size of data we used spinning drives instead of solid state drives). As a result we should obtain close to full GPU utilization during the whole learning process.

\subsection{Training on Folds of Data sub-Sample}
As discussed in the previous section, each of our model (beside LSTM) was trained on five folds of data sub-sample. Early in the data analysis stage, we came to conclusion that the data is well shuffled and highly homogeneous. Therefore for each fold we simply selected every fifth file for validation and used the other four for training. By sub-sampling the data into folds, we introduced further randomization to the trained models, and were able to achieve higher per model score by averaging the folds and checkpoints.

\subsection{Boosting Network}
Boosting is a technique widely used in statistical learning. In this competition, we designed and implemented a simple boosting network. The design is as follows. The first layer is a neural network model such as MoNN3L, or the LogisticModel from the starter code. A second neural network model will then be built to train on the errors between the first layer model's prediction and the true labels. We call this second layer model as the boosting model. The final prediction is the sum of the base model prediction plus the correction from the boosting model. There was no feedback from the boosting model to the base model. 

Since the errors of the base model range in [-1, +1], we used a $\tanh$ activation function for the boosting model, and used an L2 loss function instead of the cross-entropy loss function. We hoped that the boosting model will add value in general. However, practically we saw the boostging model helped when the base model is a weaker model ({\it e.g.}, LogisticModel) but didn't see clear added value when the base model is large. Our experiments indicated that if the base model did not digest all the information from the features, then boosting can help. A practical use case is to feed complimentary features to the boost model, when the base model focus on the major features.

\section{Summary}
In this report, we described our team's models and training techniques to solving the YouTube 8M competition. We found that both video level and frame level models can achieve GAP at 0.834 level, and the ensemble can go beyond 0.841. We found that models of larger sizes performed better. For video level models of the same size, wider models seem to perform better than deeper models. This may indicate the importance of model memory. We think our experiment with data augmentation, sub-sampling, online validation helped us to achieve better performance. The boosting model idea may be worth further discussion.

\section{Source Code}
The code we used to create all the models is available under https://github.com/mpekalski/Y8M and it is released under Apache License 2.0.

\section{Acknowledgments}
The authors would like to thank Computational Biophysics group at University Pompeu Fabra for letting us use their GPU computational resources. 
We would also like to acknowledge Jose Jimenez for valuable discussion and feedback.
{\small
\bibliographystyle{ieee}
\bibliography{egbib}
}

\end{document}